\documentclass[10pt,twocolumn,letterpaper]{article}

\usepackage{wacv}
\usepackage{times}
\usepackage{epsfig}
\usepackage{graphicx}
\usepackage{amsmath}
\usepackage{amssymb}
\usepackage{multirow}
\usepackage{amssymb}
\usepackage{float}
\DeclareMathOperator{\argmin}{argmin}
\DeclareMathOperator{\softmax}{softmax}
\usepackage{authblk}
\usepackage{color}
\usepackage[pagebackref=true,breaklinks=true,letterpaper=true,colorlinks,bookmarks=false]{hyperref}

\def\wacvPaperID{599} 

\wacvfinalcopy 

\ifwacvfinal
\def\assignedStartPage{1} 
\fi

\ifwacvfinal
\usepackage[breaklinks=true,bookmarks=false]{hyperref}
\else
\usepackage[pagebackref=true,breaklinks=true,colorlinks,bookmarks=false]{hyperref}
\fi

\ifwacvfinal
\else
\fi

\begin{document}

\title{Selective Spatio-Temporal Aggregation Based Pose Refinement System: Towards Understanding Human Activities in Real-World Videos}

\author{Di Yang\textsuperscript{1,2}\hskip 1em
Rui Dai\textsuperscript{1,2}\hskip 1em
Yaohui Wang\textsuperscript{1,2}\hskip 1em

Rupayan Mallick\textsuperscript{1}\hskip 1em
Luca Minciullo\textsuperscript{3} \hskip 1em
Gianpiero Francesca\textsuperscript{3}\hskip 1em
François Brémond\textsuperscript{1,2}
\\
\textsuperscript{1}Inria \hskip 1em
\textsuperscript{2}Université Côte d'Azur \hskip 1em
\textsuperscript{3}Toyota Motor Europe \\


{\tt\small \{di.yang, rui.dai, yaohui.wang, rupayan.mallick, francois.bremond\}@inria.fr} \hskip 1em
{\tt\small \{luca.minciullo, gianpiero.francesca\}@toyota-europe.com}

}

\renewcommand\Authands{ and }


\twocolumn[{%
\maketitle
\vspace{-1.5cm}
\begin{figure}[H]
\hsize=\textwidth 
\centering
\includegraphics[width=1.0\textwidth ]{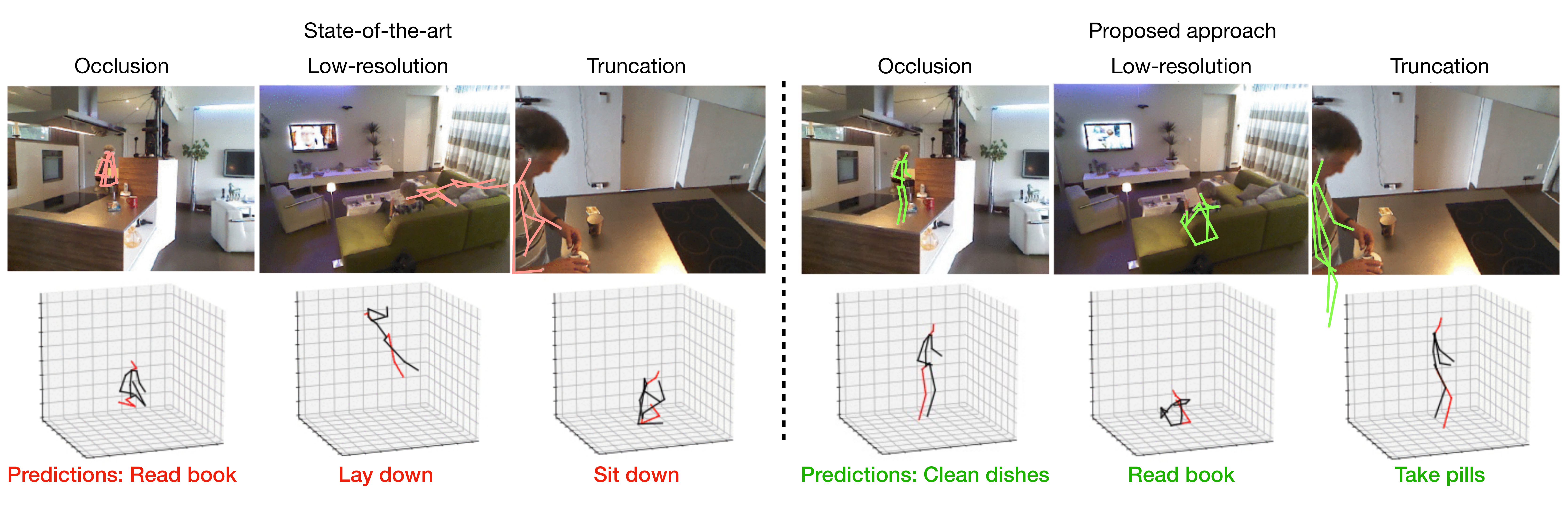}

\caption{\textbf{Skeleton-based action recognition} on Toyota Smarthome with poses (Left) extracted by AlphaPose~\cite{fang2017rmpe} (left),  LCRNet++~\cite{RogezWS18} (middle), OpenPose~\cite{OpenPose} (right) and high-quality poses (Right) obtained by the proposed pose refinement system. The action predictions using refined poses with the same action recognition system become more accurate. (3D reconstructions are from VideoPose~\cite{pavllo:videopose3d:2019} over 2D)}
\label{figure:intro}
\end{figure}
}]

\begin{abstract}
\vspace{-0.2cm}
Taking advantage of human pose data for understanding human activities has attracted much attention these days. However, state-of-the-art pose estimators struggle in obtaining high-quality 2D or 3D pose data due to occlusion, truncation and low-resolution in real-world un-annotated videos. Hence, in this work, we propose 1) a Selective Spatio-Temporal Aggregation mechanism, named SST-A, that refines and smooths the keypoint locations extracted by multiple expert pose estimators, 2) an effective weakly-supervised self-training framework which leverages the aggregated poses as pseudo ground-truth instead of handcrafted annotations for real-world pose estimation. Extensive experiments are conducted for evaluating not only the upstream pose refinement but also the downstream action recognition performance on four datasets, Toyota Smarthome, NTU-RGB+D, Charades, and Kinetics-50. We demonstrate that the skeleton data refined by our Pose-Refinement system (SSTA-PRS) is effective at boosting various existing action recognition models, which achieves competitive or state-of-the-art performance. 

\vspace{-0.5cm}
\end{abstract}


\section{Introduction}

Action recognition approaches have led to significant improvements for many applications, such as video surveillance, video understanding, human-computer interaction and game control. 
Compared with RGB-based action recognition using spatio-temporal deep convolution on RGB videos~\cite{Simonyan2014TwoStreamCN, Carreira_2017_CVPR, Wang2018TemporalHF, Ryoo2020AssembleNetAM}, skeleton-based approaches have drawn increasing attention owing to their strong ability in summarizing human motion~\cite{Vemulapalli2014HumanAR, zhang2019view, Tanfous2019SparseCO, 8784961, Xie2018MemoryAN, Li2018CooccurrenceFL, Caetano2019SkeletonIR, Yan2018SpatialTG, Li_2019_CVPR,2sagcn2019cvpr,Gao2019OptimizedSA, shi_skeleton-based_2019, Shi2019SkeletonBasedAR, liu2020disentangling}. As high level representations, skeletons have the merits of being robust to changes of appearances, environments, and view-points. Earlier skeleton-based approaches using deep-learning like RNNs~\cite{zhang2019view, Tanfous2019SparseCO, 8784961, Song2017AnES, Xie2018MemoryAN} or temporal CNNs~\cite{Li2018CooccurrenceFL, Caetano2019SkeletonIR} are proposed owing to their high representation capacity, but they ignore the important semantic connectivity of the human body. Recent GCNs-based approaches~\cite{Yan2018SpatialTG, Li_2019_CVPR,2sagcn2019cvpr,Gao2019OptimizedSA, shi_skeleton-based_2019, Shi2019SkeletonBasedAR, liu2020disentangling} construct spatial-temporal graphs and model the spatial relationships within GCNs directly, and these methods have seen significant performance boost, indicating the interest of using semantic human skeleton for action recognition. However, most of these algorithms use accurate 3D human poses obtained from a Motion Capture system~\cite{Shahroudy2016NTURA} and they cannot perform well on real-world videos with low-quality 2D or 3D poses. In the real-world, 1)~the performance of human pose estimation approaches~\cite{RogezWS18, He_2017_ICCV, OpenPose, fang2017rmpe, Kreiss_2019_CVPR, Wang2018MagnifyNetFM, Gler2018DensePoseDH} is limited, especially in case of occlusion, truncation, and low-resolution scenarios. 2)~A large amount of manual skeleton annotation is extremely expensive to obtain. Motivated by the above problem, we propose a skeleton-based action recognition framework that uses a Selective Spatio-Temporal Aggregation based Pose Refinement System, named SSTA-PRS, to \textbf{extract high-quality 2D skeletons from un-annotated real-world videos} and leverages AGCNs~\cite{2sagcn2019cvpr} for action classification and action detection. \par

To deal with \textbf{the absence of keypoints (\ie joints) due to occlusion, truncation and low-resolution} in real-world pose estimation (as shown in Fig.~\ref{figure:intro}), we construct 
a multi-expert pose estimation system to predict improved 2D poses. It is fine-tuned with the pseudo ground-truth 2D pose generated by a novel Selective Spatio-Temporal Aggregation (SST-A) which integrates the pose proposals computed from several existing expert pose estimators. In this work, we select LCRNet++~\cite{RogezWS18}, OpenPose~\cite{OpenPose} and AlphaPose~\cite{fang2017rmpe} as the experts. 
Regarding on our downstream tasks, we leverage the AGCNs~\cite{2sagcn2019cvpr} to extract features for an input pose sequence that are used directly for action classification (for trimmed videos) or fed into a temporal model like LSTM~\cite{graves2005framewise}, TCNs~\cite{lea2017temporal} for temporal action detection (for untrimmed videos). The contributions related to this paper are summarized as follows:
\begin{itemize}
\item We propose a novel Selective Spatio-Temporal Aggregation mechanism (SST-A), that integrates the advantage of several expert pose estimation systems in both spatial and temporal domains, and we introduce a confidence metric $C$ to evaluate the quality of the aggregated poses. 
\item We present a weakly-supervised self-training Pose Refinement System (SSAT-PRS) based on LCRNet++~\cite{RogezWS18} using pseudo-ground truth poses, generated by our SST-A mechanism instead of using handcrafted pose annotations. 
\item We directly evaluate the upstream pose refinement method using pose ground-truth then demonstrate the effectiveness of this framework for the downstream GCNs-based action recognition task using action ground-truth. 
\end{itemize}
Experiments are conducted on four real-world datasets, Toyota Smarthome, NTU-Pose, Charades and Kinetics-50.

\section{Related Work}

\textbf{Human Pose Estimation in Real-World.}
Most state-of-the-art approaches for 2D human pose estimation employ 2D CNNs architectures for a single image in a strongly-supervised setting~\cite{ning2017knowledge, OpenPose, Wang2018MagnifyNetFM, fang2017rmpe,He_2017_ICCV,cheng2020bottom, Kreiss_2019_CVPR, multitask}. For 3D pose estimation, \cite{RogezWS18, Moon_2019_ICCV_3DMPPE} focus on end-to-end reconstruction by directly estimating 3D poses from RGB images without intermediate supervision. \cite{zhaoCVPR19semantic} applies GCNs for regression tasks, especially 2D to 3D human pose regression. \cite{pavllo:videopose3d:2019} demonstrates that 3D poses in video can be effectively estimated with a fully convolutional model based on dilated TCNs over 2D keypoint sequences. Among these methods, \cite{RogezWS18,ning2017knowledge, Moon_2019_ICCV_3DMPPE, fang2017rmpe, He_2017_ICCV,Wang2018MagnifyNetFM} have first to incorporate a person detector, followed by the estimation of the joints and then the computation of the pose for each person. These approaches give full-body predictions once the people are detected, but the detection speed slows down with the increase of the number of people present in the image. \cite{OpenPose, cheng2020bottom,Kreiss_2019_CVPR, multitask} are bottom-up approaches which detect all joints in the image using heatmaps that estimate the probability of each pixel to correspond to a particular joint, followed by associating body parts belonging to distinct individuals. These approaches cannot always provide the none-visible body parts for each individual due to occlusions and truncations. \par
 
By annotating poses in the real-world, approaches~\cite{RogezWS18, OpenPose, fang2017rmpe, rockwell2020fullbody} are becoming more robust to occlusion and they can provide us with  pre-trained pose estimators, so that we can extract skeleton data from real-world videos without expensive handcraft annotations.
In particular, LCRNet++~\cite{RogezWS18} is an attractive pose estimator, which leverages a Faster R-CNN~\cite{renNIPS15fasterrcnn} like architecture with a CNNs backbone. A Region Proposal Network extracts candidate boxes around humans. To deal with occlusions and truncation, LCRNet++ proposes ‘anchor-poses’ for pose classes instead of object classes: 
these key poses typically correspond to a person standing, sitting, etc. Bottom-up method OpenPose~\cite{OpenPose} proposes an alternative approach by regressing affinities between joints (\ie the direction of the bones), together with the heatmaps. AlphaPose~\cite{fang2017rmpe} improves the performance of top-down pose estimation algorithms by detecting accurate human poses even with inaccurate bounding boxes. 
Closer to our work, Rockwell et al.~\cite{rockwell2020fullbody} propose an effective self-training framework that adapts human 3D mesh recovery systems to consumer
videos. They focus on recovering from occlusions and truncations, but they do not have solutions to tackle low-resolution images and the instability of the extracted 3D meshes along time. In our work, we combine the advantages of the three expert pose estimators~\cite{RogezWS18, OpenPose, fang2017rmpe} by spatio-temporal aggregating their results and getting a more accurate pose than using only one expert.\par

 
\begin{figure*}
\begin{center}

\includegraphics[width=1.0\linewidth]{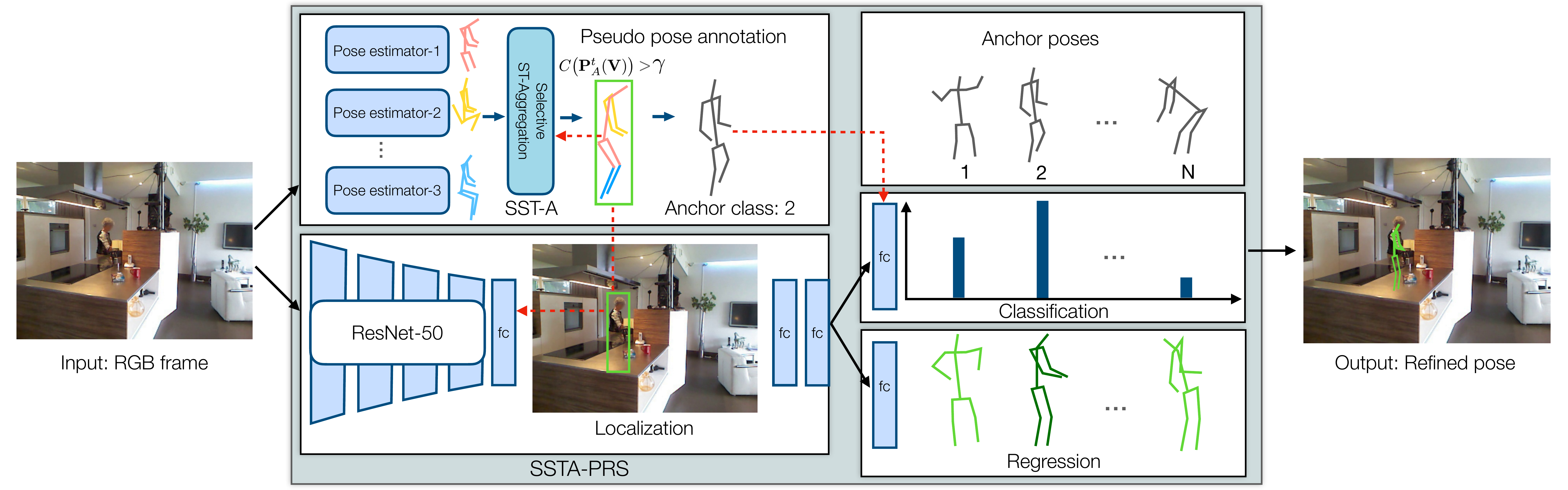}
\end{center}
   \vspace{-0.2cm}
   \caption{\textbf{Overview of Pose-Refinement System (SSTA-PRS).} Given an RGB frame, a less noisy 2D pose $\mathbf{P}^t_A\mathbf{(V)}$ and its confidence value $C$ is computed by the Selective Spatio-Temporal Aggregation (SST-A) with the pose proposals obtained by several pose estimation systems and the previous aggregated pose. 
   If the confidence is higher than the threshold $\gamma$, we are able to calculate the pseudo ground-truth bounding box and anchor class according to this improved pose to fine-tune the localization and classification branches of an LCRNet~\cite{RogezWS18} architecture in a weakly-supervised setting. Finally, this refined pose estimation system is used to extract high-quality 2D poses in the real-world videos for the downstream action recognition task.}
\vspace{-0.2cm}
\label{fig:prs}
\end{figure*}


\textbf{Skeleton-Based Action Classification.}
ST-GCN~\cite{Yan2018SpatialTG}, where spatial graph convolutions along with interleaving temporal convolutions are used for spatial-temporal modeling is the first GCNs-based method for action recognition. Subsequently, many GCNs methods representing a more appropriate spatial graph on intra-frames have been proposed.
2s-AGCN~\cite{2sagcn2019cvpr} introduces an adaptive graph convolutional network to adaptively learn the topology of the graph with self-attention, which can better suit the action recognition task and the hierarchical structure of the GCNs. It also uses a two-stream ensemble with skeleton bone features to boost performance. Their further work, MS-AAGCN~\cite{shi_skeleton-based_2019} proposes multi-stream adaptive graph convolutional networks that introduce attention modules and a multi-stream ensemble based on 2s-AGCN~\cite{2sagcn2019cvpr}. Note that these approaches primarily focus on spatial modeling. In contrast, MS-G3D~\cite{liu2020disentangling} presents a unified approach for capturing complex joint correlations directly across spacetime. \par
Yan et al.~\cite{Yan2018SpatialTG} use OpenPose~\cite{OpenPose} to extract poses for applying GCNs on the real-world dataset~\cite{Carreira_2017_CVPR}, but the pose data is not always full-body due to occlusions and truncations.\par
\textbf{Skeleton-Based Action Detection.}
Action detection (\ie action localization) task needs to find precise temporal boundaries of actions occurring in an untrimmed video. 
In order to model long temporal relationships, the current detection methods~\cite{lea2017temporal,TGM,superevent} encode the videos as a pre-processing step. 
After that, action detection can be seen as a sequence-to-sequence problem.
Recurrent neural networks (RNNs)~\cite{LSTM_basic,multi-thumos} have been popularly used to model actions transitions between frames. 
These approaches relied on the memory cell to accumulate temporal information in different frames, thus implicitly capturing the relationships between actions.
Temporal convolution networks (TCNs) are another type of action detection model.
Dilated-TCN~\cite{lea2017temporal} increases the temporal reception field by using dilated convolutions to model long temporal patterns. 
Temporal Gaussian Mixture~\cite{TGM} utilizes the Gaussian distribution to generate the weights of the temporal kernel. This Gaussian-based kernel enables TGM to capture long temporal relations via a large receptive field with limited parameters. 

In our work, we apply AGCNs~\cite{2sagcn2019cvpr} that performs the best on Smarthome~\cite{Das_2019_ICCV} with estimated and refined pose data instead of using a Motion Capture system. To the best of our knowledge, we are the first to provide high-quality full-body poses using a pose refinement system and leverage AGCNs~\cite{2sagcn2019cvpr} for both action classification and detection in real-world datasets~\cite{Das_2019_ICCV, Sigurdsson2016HollywoodIH} without ground-truth poses. \par


\section{Pose Refinement Approach}

The overall architecture of the proposed method for pose refinement is shown in Fig.~\ref{fig:prs}. Given an RGB frame, several pose proposals are obtained by multiple expert pose estimation systems~\cite{RogezWS18, He_2017_ICCV, OpenPose, fang2017rmpe, Kreiss_2019_CVPR, Wang2018MagnifyNetFM, Gler2018DensePoseDH} and the Selective Spatio-Temporal Aggregation mechanism (SST-A) computes an improved pose, which is more accurate, smoother, and more stable along time. 
With this aggregated pose, we compute a confidence metric to estimate its quality. Then, we select the aggregated poses with higher confidences than a threshold and calculate the pseudo ground-truth bounding box and anchor class to fine-tune the localization and classification branches of an LCRNet~\cite{RogezWS18} architecture.~
Finally, the refined pose estimation system is used to extract higher-quality poses in real-world videos.\par

\subsection{Selective Spatio-Temporal Aggregation} \label{ssta}
Selective Spatio-Temporal Aggregation (SST-A) is the key component to deal with the \textbf{absence of keypoints} caused by occlusion, truncation and low-resolution, and with the \textbf{instability in time domain} due to pose estimation from a single frame. 
Our insight is that 1) bottom-up methods directly predict the keypoints through the heatmaps, however, they may miss joints that are none-visible due to occlusions or truncations because the number of each body part prediction may not correspond to the number of people in the image. 2) Top-down methods regress the coordinates of keypoints over the bounding box of the people. As long as people are detected, the keypoints of full-body can be predicted. But these methods may miss people in low-resolution images, resulting in missing all the joints of these people. According to the above analysis, by combining the results of both families of methods, we can reduce the number of missing joints and obtain more stable and higher-quality full-body keypoints. Therefore, we leverage multiple expert pose estimation systems, including methods from both families to extract poses for the same frame as several pose proposals, and then aggregate them to recover the missing keypoints. In this work, we select two top-down estimation systems LCRNet++~\cite{RogezWS18} and AlphaPose~\cite{fang2017rmpe} and a bottom-up estimation system OpenPose~\cite{OpenPose} to provide the pose proposals,
which are then combined into an improved pose sequence through our SST-A mechanism. Moreover, our pose sequence is extracted frame by frame with the estimators, so there is a certain lack of temporal continuity, resulting in a static joint shaking in the video. This problem is also an obstacle for the performance of action recognition. Hence, our SST-A also uses a temporal filtering mechanism to smooth the entire sequence by eliminating unstable values. \par

\begin{figure}[t]
\begin{center}
   \includegraphics[width=0.9\linewidth]{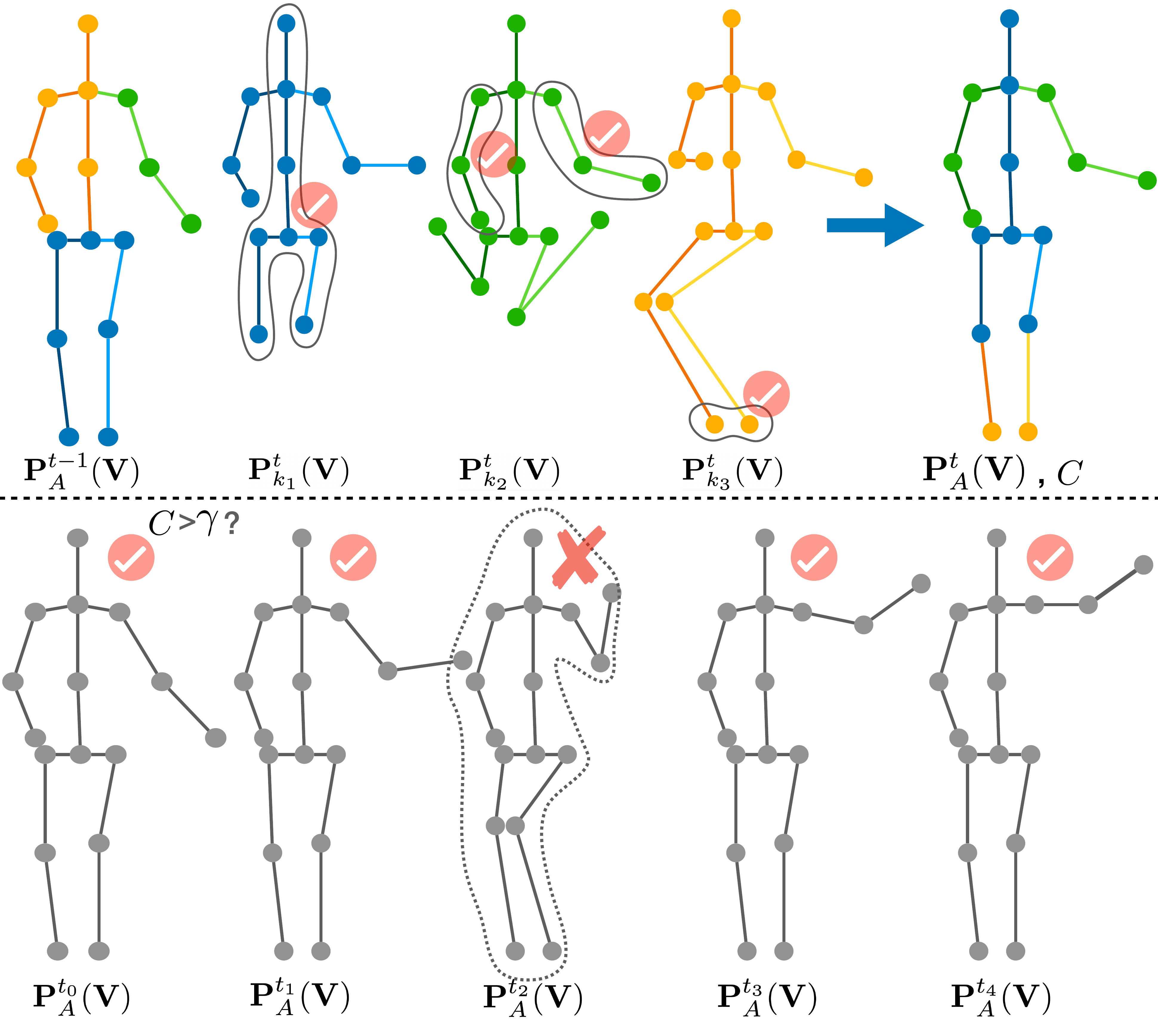}
   
\end{center}
\vspace{-0.4cm}
   \caption{\textbf{Two steps of SST-Aggregation}. 1) Aggregation in both spatial and temporal level (top). The pose $\mathbf{P}^t_A\mathbf{(V)}$ of current frame is aggregated from the three pose proposals $\mathbf{P}^t_{k_1}\mathbf{(V)}$ (blue), $\mathbf{P}^t_{k_2}\mathbf{(V)}$ (green) and $\mathbf{P}^t_{k_3}\mathbf{(V)}$ (yellow). 2) Selective temporal filter (bottom). The pose with a low confidence in the aggregated sequence will be discarded.}
\vspace{-0.3cm}
\label{fig:sta}
\end{figure}

As shown in Fig.~\ref{fig:sta}, we note all the $N$ keypoints in one body as a set $\mathbf{V=\{ v_1, v_2, ... , v_N\}}$, the frames as a set $F=\{ 0, 1, 2, ... , T \}$ and the position of one joint $\mathbf{v} \ (\mathbf{v} \in \mathbf{V})$ in the frame $t \ (t \in F)$ estimated by the pose estimation system $k_m \ (k_m \in K=\{k_1, k_2, ..., k_M \})$ as $\mathbf{P}^t_{k_m}\mathbf{(v)}$, noted that $K$ is the ensemble of pose estimation systems. The final aggregated pose sequence of the body $\mathbf{V}$ is noted as $\mathbf{P}^F_A\mathbf{(V)}=\{\mathbf{P}^t_A\mathbf{(v)} | \mathbf{v} \in \mathbf{V}, t \in F \}$ and our aggregation system has two steps. \par

\textbf{1) Joint-level aggregation.} Each keypoint of the pose is calculated from the prediction results of the three estimators~\cite{OpenPose,RogezWS18,fang2017rmpe} in the current frame $\mathbf{P}^t_{k_m}\mathbf{(v)}$ and the aggregated result of the previous frame $\mathbf{P}^{t-1}_A\mathbf{(v)}$. So this step is to select the closest keypoint to the same part aggregated in the previous frame. For the first frame, we select any one of the two closest keypoints obtained by pose estimators. Joint-level aggregation can be written as:
\begin{small}
\begin{equation}\label{sst2}
    \mathbf{P}^t_A(\mathbf{v})=
    \begin{cases}
        \mathbf{P}^t_{k_a}\mathbf{(v)},~\text{if}~ t=0\\
        (k_a, k_b)=\underset{(k_i, k_j) \in K^2, i \neq j}{\argmin}\Big (D \big (\mathbf{P}^t_{k_i}\mathbf{(v)}, \mathbf{P}^t_{k_j}\mathbf{(v) \big)} \Big) \\ 
        \mathbf{P}^t_{k_a}\mathbf{(v)},~\text{if}~t>0 \\
        k_a=\underset{ k_i \in K}{\argmin} \Big( D\big(\mathbf{P}^t_{k_i}\mathbf{(v)}, \mathbf{P}^{t-1}_A\mathbf{(v) \big)  } \Big) \\
    \end{cases}
\end{equation}
\end{small}
where $D$ is the Euclidean distance between two key-points in the image, noted as: 
\begin{small}
\begin{equation}\label{sst3}
    D\big(\mathbf{P}_{1}\mathbf{(v)},\mathbf{P}_{2}\mathbf{(v)}\big)= \sqrt{\big(\mathbf{P}_{1}\mathbf{(v)}-\mathbf{P}_{2}\mathbf{(v)}\big)^{2}}
\end{equation}
\end{small}
\textbf{2) Body-level aggregation.} Followed by the first step which can effectively solve the problem of missing keypoints, we define a \textbf{confidence metric $C \in (0, 1]$} that describes the likelihood that the aggregated pose is the real pose in order to further smooth the pose sequence. We believe that when the average similarity between the aggregated pose and the pose proposals is very high, the pose proposals are also very similar, indicating that the pose proposal itself is likely to be accurate, and the aggregation result will have a high confidence. This selective likelihood filter is written as~\eqref{sst4}, which is to discard the abnormal poses with a very low confidence in the whole sequence,
\begin{small}
\begin{equation}\label{sst4}
    \mathbf{P}^t_A\mathbf{(V)}=
    \begin{cases}
    \mathbf{P}^t_A\mathbf{(V)},~\text{if}~C\big(\mathbf{P}^t_A\mathbf{(V)}\big)>=\gamma\\ 
    \text{discard},~\text{if}~C\big(\mathbf{P}^t_A\mathbf{(V)}\big)<\gamma
    \end{cases}
\end{equation}
\end{small}
where $C$~\eqref{sst5} is defined to describe the confidence of this aggregated pose. ($D_{normal}$ is the distance between the aggregated head and neck while offset $\epsilon=10^{-12}$ is to prevent errors in case of $D_{normal}=0$)
\begin{small}
\begin{equation}\label{sst5}
    C\big(\mathbf{P}^t_A\mathbf{(V)}\big)= \exp\Big(-\cfrac {1}{NM} \sum_{\mathbf{V}}\sum_{K} \cfrac{D\big(\mathbf{P}^t_A\mathbf{(v)},\mathbf{P}^t_{k_m} {\mathbf{(v) \big) }}}{D_{normal}+\epsilon}\Big)
\vspace{-0.2cm}
\end{equation}
\end{small}

$\gamma$ is a filtering parameter that represents a threshold. If the confidence of the pose in the current frame is lower than this threshold, it will be discarded from the sequence. After this two-step SST-A is completed, we obtain a higher-quality full-body skeleton sequence from a video, which can be effectively used as pseudo ground-truth pose for our self-training Pose Refinement system in Sec.~\ref{st}.\par

\subsection{Self-Training Pose Refinement System} \label{st}

SST-A can effectively integrate the advantages of~\cite{RogezWS18, OpenPose, fang2017rmpe}. However, this aggregation method may increase the workload in practice because we have to estimate the poses several times with different systems. Hence, we propose a self-training framework using the higher-quality 2D poses obtained from SST-A as supervised pseudo ground-truth, to refine one of the pose estimation models. Once the model is refined, the other models are not needed for inference. In fact, we only need to run SST-A on a small part of the dataset, and then it can be used for fine-tuning the network. As shown in Fig.~\ref{fig:prs}, we build our pose refinement model (SSTA-PRS) based on LCRNet++~\cite{RogezWS18} owing to its particularity of its three branches (localization, classification and regression), we do not have to provide truly accurate pose labels but only fine-tune the localization and classification branches with the pseudo ground-truth 2D poses.\par
\vspace{-0.2cm}
\subsubsection{Overview of SSTA-PRS architecture}
As LCRNet++~\cite{RogezWS18}, our SSTA-PRS framework also contains 4 main components. 1) \textbf{Localization:} it leverages a Faster R-CNN~\cite{renNIPS15fasterrcnn} like architecture with a ResNet-50 backbone~\cite{He_2016_CVPR}. Given an input image, a Region Proposal Network (RPN)~\cite{renNIPS15fasterrcnn} extracts candidate boxes around humans. 2) \textbf{Classification:} these regions are then classified into different ‘anchor-poses’ pre-defined by K-means clustering that typically correspond to a person standing, a person sitting, etc. In this paper, 'anchor-poses' are defined in 2D only, and the refinement occurs in this joint 2D pose space. 3) \textbf{Regression:} a class-specific regression is applied to estimate body joints in 2D. First, for each class of pose, we define offline the ‘anchor-poses’, computed as the center over all elements in the corresponding cluster. After fitting all the 2D anchor-poses into each of the candidate boxes, we perform class-specific regressions to deform these anchor-poses and match the actual 2D pose in each box. 4) \textbf{Post-processing:} for each individual, multiple pose candidates can overlap and produce valid predictions. These pose candidates are combined by pose proposal integration~\cite{RogezWS18}, taking into account their 2D overlap and classification scores. As the approach is holistic, it outputs full-body poses, even in case of occlusions or truncation by image boundaries.\par
\vspace{-0.2cm}
\subsubsection{Weakly-supervised training}
We train this model with a weakly-supervised setting, which only refines the 2D localization and classification. The reason is that firstly, our pseudo pose annotations are not sufficiently accurate for regression while they are accurate enough for localization and classification. Secondly, the in-the-wild pre-trained model has good prediction performance when the localization and classification are correct. However, in low-resolution images, the bounding boxes of people are usually very difficult to search, which may result in no estimated keypoint on the body, or an error in the classification stage leading to inaccurate pose prediction. Therefore, if the classification and localization branches are correctly fine-tuned, the model should find the correct anchor class so that the final prediction can be more accurate.\par
\textbf{Pseudo 2D pose ground-truth:} it contains two parts, the bounding box of people and the anchor class. Both are calculated using the SST-A pose results. We take the maximum and minimum values of the pose in $x$ and $y$ directions as the boundary of the initial bounding box. We then expand the box by $10\%$ as ground-truth for the localization branch, because the key-points do not correspond exactly to the boundary of the person. The class label of pose $P$, noted as $Class_P \in \{0,1,...,B\}$, is set by finding the closest 2D anchor-pose $Anchor_P$ according to the similarity $S$~\cite{RogezWS18} between the oriented 2D poses centered at the left-top corner of bounding box: $Class_P = \argmin_b S(Anchor_b, P)$. This label is used by the classification branch as pseudo ground-truth.\par
\textbf{Loss function:} our loss is the sum of the following two losses, described as:
\begin{equation}\label{loss1}
    L=L_{loc}+L_{classif}
\end{equation}
The loss of the localization component is the loss of the region proposal network~\cite{renNIPS15fasterrcnn} (RPN):
\begin{equation}\label{loss2}
    L_{loc}=L_{RPN}
\end{equation}
Same as~\cite{RogezWS18}, let $u$ be the probability distribution estimated by SSTA-PRS, obtained by the fully connected layers of the classification branch after RoI pooling, followed by a $\softmax$ function. The classification loss is defined using the standard cross entropy loss:
\begin{equation}\label{loss3}
    L_{classif}(u,Class_P)=-\log u(Class_P)
\end{equation}
\par

\section{Experimental Setup}
\begin{figure*}[ht]
\begin{center}
\includegraphics[width=1.0\linewidth]{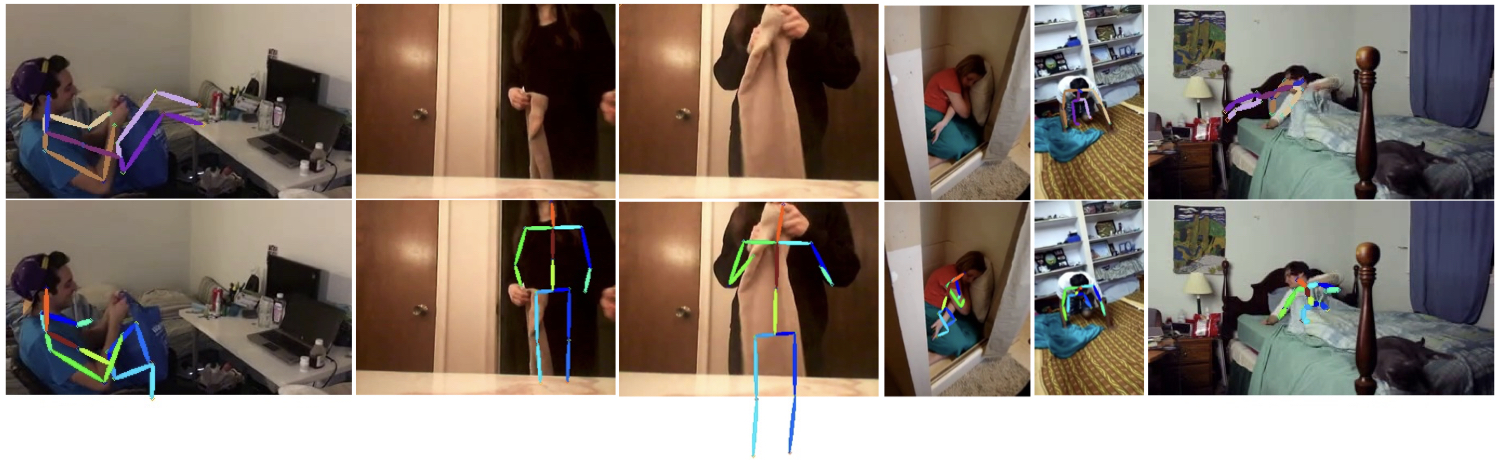}
\vspace{-0.8cm}
\end{center}
   \caption{Comparison of the poses extracted on Charades by LCRNet++~\cite{RogezWS18} without refinement (top) with our SSTA-PRS (bottom).}
\vspace{-0.2cm}
\label{fig:charades}
\end{figure*}

Our objective is to obtain high-quality poses from real-world videos in order to understand human activities. We conduct a wealth of experiments to evaluate our system with two protocols: 1) Evaluation by the upstream pose refinement task using pose ground-truth, which is to directly compare the accuracy of the poses obtained from the pose estimators~\cite{RogezWS18, OpenPose, fang2017rmpe} with our proposed SSTA-PRS. 2) Evaluation by the downstream action recognition task using ground-truth action labels. We use the same action recognition model~\cite{2sagcn2019cvpr}, but processing different pose data with and without refinement for comparing the action recognition (classification and detection) performances, which indirectly demonstrates the effectiveness of our SSTA-PRS.

\subsection{Datasets and Evaluation Protocols}
\textbf{Toyota Smarthome} (Smarthome)~\cite{Das_2019_ICCV} is a recent real-world daily living dataset for action classification, recorded in an apartment where 18 older subjects carry out tasks of daily living during a day. The dataset contains 16,115 videos of 31 action classes, and the videos are taken from 7 different camera viewpoints. All the actions are performed in a natural way without strong prior instructions. This dataset provides RGB data and 3D skeletons which are extracted from LCRNet++~\cite{RogezWS18}. In contrast, in our experiments, we extract higher-quality 2D skeleton data with our SSTA-PRS framework instead of the provided ones. For the evaluation on this dataset, we follow cross-subject (CS) and cross-view (CV1 and CV2) protocols. We use Smarthome as our main dataset for ablations. 
\par

\textbf{Smarthome-Pose:} in order to evaluate directly the poses extracted by our SSTA-PRS,  we chose the middle frames for randomly selected 1,400 videos of Toyota Smarthome and we annotated the 2D poses to create a test set containing 1,400 images with $640\times480$ resolution and many occlusions, truncations. We follow the PCKh $@0.5$ (percent of keypoints within a threshold of $0.5$ times head length) as the pose evaluation protocol. We regard the distance in pixels between head and neck as the head length.\par

\textbf{NTU-Pose:} NTU-RGB+D~\cite{Shahroudy2016NTURA} is a large-scale multi-modal dataset which consists of 56,880 sequences of high-quality 2D/3D skeletons with 25 joints, associated with depth maps, RGB and IR frames captured by the Microsoft Kinect v2 sensor. We selected 60 videos (6,098 frames) with the same subject performing different actions and we took the 2D skeleton as the ground-truth for pose evaluation. The dataset was recorded in a laboratory, so in this work, we changed the original quality of the videos by reducing the resolution to $320\times180$ and adding partial occlusions to make it similar to our real-world settings. We use the 2D MPJPE (mean per joint position error normalized by head length) and PCKh $@2.0$ protocols. \par


\textbf{Charades}~\cite{Sigurdsson2016HollywoodIH} is a challenging dataset due to large environmental diversity. 
The videos were recorded in the homes of 267 subjects. 
While performing the actions, the subjects are usually occluded by objects or the furniture of the scene. 
%
We apply SSTA-PRS on Charades to validate whether this algorithm can extract robust skeletons even when subjects are occluded. (Fig.~\ref{fig:charades})
Note that there are 157 action classes in the Charades dataset and most of them are relevant to objects but not to poses. As we only use the skeletons to recognize actions, we kept the 31 meaningful semantic verbs as the action classes. These semantic verbs and additional results using the original class labels are provided in supplementary material.  
%
We follow the localization setting of the dataset, reporting the frame-based mAP. \par

\textbf{Kinetics-50} is a subset of a large-scale real-world dataset Kinetics-400~\cite{Carreira_2017_CVPR} that contains about 300,000 video clips for 400 action classes collected from YouTube and does not contain pose information. 
We took the first 50 action categories in alphabetical order corresponding to 27,314 video clips and we extracted the 2D poses with our SSTA-PRS module. Following the evaluation method in~\cite{Yan2018SpatialTG}, we train the action recognition model on the training set (25,288 video clips) and report the top-1 and top-5 accuracy on the validation set (2,026 video clips).

\subsection{Implementation Details}\label{imp}

\textbf{Pose estimators:} we select 1)~OpenPose 18-joints~\cite{OpenPose}, 2)~AlphaPose Sample-Baseline~\cite{fang2017rmpe} using YOLOv3~\cite{Redmon2018YOLOv3AI} as detector and 3)~LCRNet++ In-The-Wild~\cite{RogezWS18} as three expert pose estimation models and LCRNet++ In-The-Wild~\cite{RogezWS18} as the student model for refining the pose. The poses contain the 13 main joints that all three estimators can detect.\par

\textbf{Pose refinement:} we select all the videos from NTU-Pose, $10\%$ of the videos from Smarthome and $9.0K$ videos from Charades, and $10\%$ of the frames for each video with uniform sampling to get a large dataset containing $3.0K$ images from NTU-Pose, $40.2K$ images from Smarthome, and $65.9K$ from Charades. We then split $20\%$ of the images as the validation set. We apply SST-A mechanism using~\cite{RogezWS18, fang2017rmpe, OpenPose} as ${k_1}, {k_2}$ and ${k_3}$ and take 13 main keypoints for aggregation with $\gamma=0.18$ as the confidence threshold for temporal filter to generate pseudo ground-truth 2D poses (Sec.~\ref{ssta}). Then, we use In-The-Wild pre-trained model of LCRNet++~\cite{RogezWS18}, which sets $20$ 'anchor poses' and leverages ResNet50~\cite{He_2016_CVPR} as a backbone and we follow the standard setting values from~\cite{RogezWS18}. We fine-tune this model using $4$ images per batch, and $512$ boxes per image. The refined model is used to estimate 2D poses of the whole set of Smarthome, Charades and  also for Kinetics-50 although its videos are not in the training set. \par

\section{Evaluation Using Pose Ground-Truth}
To evaluate the performance of our upstream pose refinement system (SSTA-PRS), we experiment on NTU-Pose and Smarthome-Pose to directly compare the performance of expert pose estimators with our SSTA-PRS.
\subsection{Results And Discussion}
\textbf{SST-A:} We estimate 2D poses using three expert estimators and then perform SST-A without discarding any frames. The results in Tab.~\ref{tab:pose1} show that SST-A is effective to integrate the advantages of the expert estimators and achieves a better performance ($+7.7\%$ on NTU-Pose, $+1.3\%$ on Smarthome-Pose). \par
\begin{figure}[ht]
\begin{center}
   \includegraphics[width=1.0\linewidth]{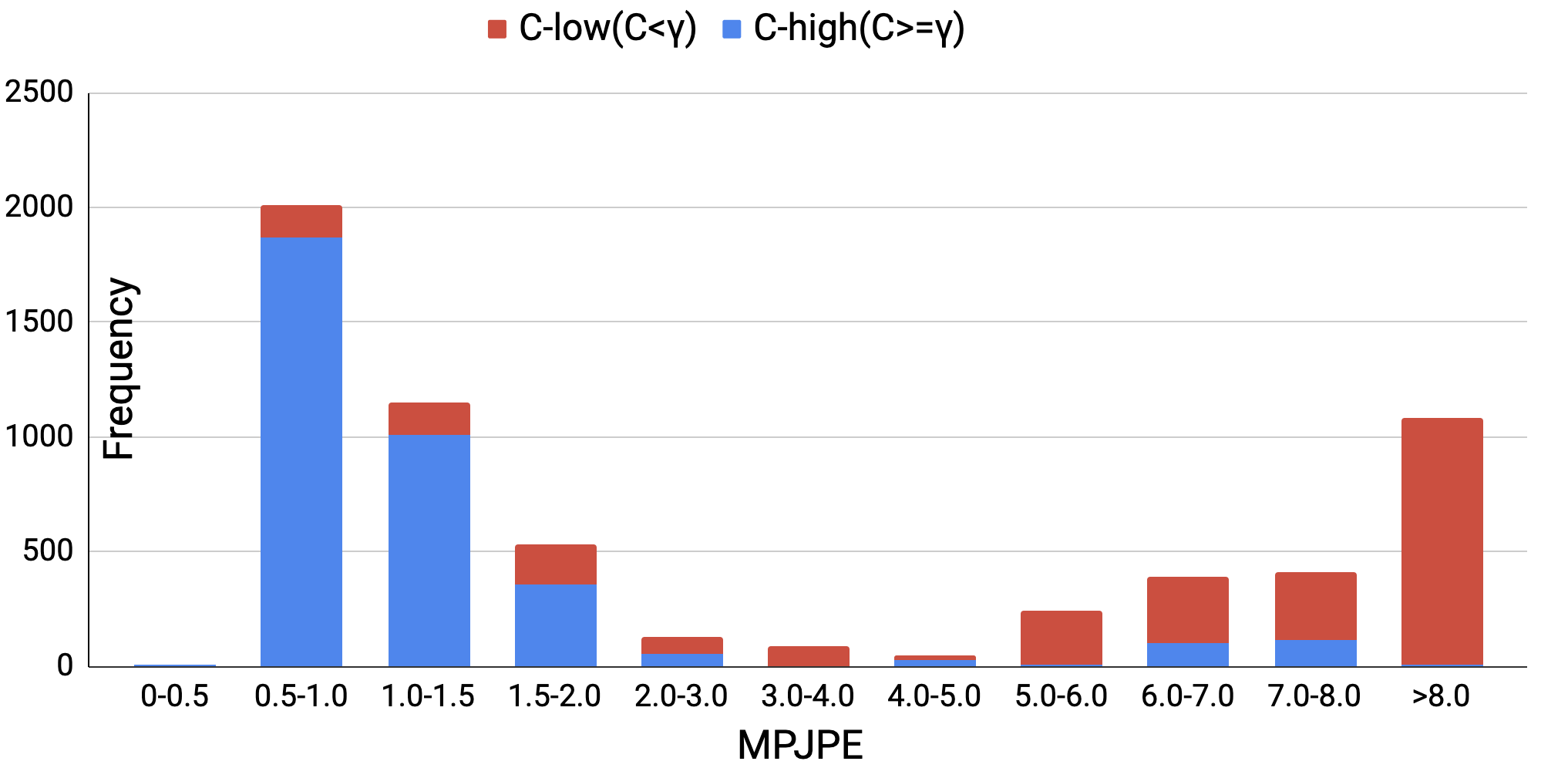}
\end{center}
    \vspace{-0.4cm}
   \caption{Histogram of pose frequency in function of MPJPE with threshold $\gamma=0.18$ (\ie high confidence when C $> \gamma$).}
   \vspace{-0.2cm}
\label{fig:exp_ssta}
\end{figure}

\textbf{Confidence metric:} To analyze the reliability of the confidence metric (Sec.~\ref{ssta}) that filters the poses for pseudo annotations, we analyse on the NTU-Pose the variation of the MPJPE with the confidence $C$ and find that the confidence decreases globally with an increase of the error. 
We select $\gamma=0.18$ as the confidence threshold that can keep most of the aggregated poses within the error of $3.0$ (Visualization in the supplementary material). According to this threshold, we analyze the frequency of the retained (\ie with C-high confidence) and discarded (\ie with C-low confidence) poses within different error intervals (Fig. \ref{fig:exp_ssta}). Within the intervals of smaller errors, we keep the most of the poses and remove the ones in the larger error intervals. In order to have sufficient training samples, 
we still keep some poses with a few errors (but high confidence), corresponding to cases of complex scenes. Our fine-tuning system is weakly-supervised training, these poses can still play a positive role in localization and classification. Therefore, the confidence metric is instrumental in our work.\par
\textbf{SSTA-PRS:} After this filtering stage, we can get higher-quality pseudo 2D pose annotations for fine-tuning SSTA-PRS. Compared with the three expert estimators (Tab.~\ref{tab:pose1}), our fine-tuned SSTA-PRS is the most effective ($+13.9\%$ on NTU-Pose and $+9.3\%$ on Smarthome-Pose).

\begin{table}[ht]
\centering

\begin{center}
\scalebox{0.8}{
\setlength{\tabcolsep}{4.0mm}{
\begin{tabular}{ l c c }
\hline
\multirow{2}*{\textbf{Methods}} &{\textbf{NTU-Pose}}& {\textbf{Smarthome-Pose}}\\
&\text{PCKh $@2.0$ (\%)} &\text{PCKh $@0.5$ (\%)}\\
\hline
\hline
LCRNet++~\cite{RogezWS18} &54.1 &  64.4\\
\text{AlphaPose~\cite{fang2017rmpe}}& \text{53.2}& 55.5 \\
\text{OpenPose~\cite{OpenPose}}& \text{45.4}& 58.9 \\
\hline
\text{SST-A only(ours)}& \text{61.8}&  65.7\\
\textbf{SSTA-PRS(ours)}& \textbf{68.0} & \textbf{73.7}\\

\hline

\end{tabular}
}}

\end{center}
\vspace{-0.2cm}
\caption{PCKh of poses from different pose estimators and proposed SSTA-PRS using SST-A only (Sec.~\ref{ssta}) and using both SST-A and self-training (Sec.~\ref{st}) on NTU-Pose and Smarthome-Pose.}
\vspace{-0.2cm}
\label{tab:pose1}
\end{table}

\section{Evaluation Using Action Ground-Truth}
We conduct several experiments on Toyota Smarthome, Charades and Kinetics-50 for the downstream action recognition task to evaluate the poses extracted through the proposed SSTA-PRS. We perform an ablation study to validate the effectiveness of our pose refinement system. Then, we compare our system with other state-of-the-art methods.
\vspace{-2mm}
\subsection{Action Recognition Approach}
We compare the performance of the best state-of-the-art action recognition model on Smarthome, AGCN~\cite{2sagcn2019cvpr}, with and without our pose refinement module.\par
\textbf{Action Classification:} Unless stated, we only use 2D poses for action recognition, which reduces the computation costs without losing performance as we have accurate enough 2D poses. We call the implemented model 2s-AGCN+SSTA-PRS. We also expand the channels of 2s-AGCN~\cite{2sagcn2019cvpr} to five as 5C-AGCN
, which concatenates the centered 3D poses to handle relative motion dynamics and 2D poses to obtain the global trajectory information.

\textbf{Action Detection:} Similar to other action detection models~\cite{graves2005framewise,TGM,lea2017temporal}, the input to the baseline models is the segment-level video encoding.
In this work, we utilize AGCNs~\cite{2sagcn2019cvpr} to encode the skeleton stream. The model is fine-tuned on the training set. 
To extract the features, a video is divided into $T$ non-overlapping segments, each segment consisting of 16 continuous poses. These segments are sent to the fine-tuned AGCNs model to extract the segment-level features. 
This video representation is the input sequence to the action detection models~\cite{graves2005framewise,TGM,lea2017temporal}.

\subsection{Ablation Study}
We analyze the impact of our two components for pose refinement. Performance is reported as mean per-class accuracy on Smarthome using only joint data.\par

\textbf{Impact of pose refinement:}
we verify the effectiveness of the proposed pose refinement system using SST-A only or both SST-A and self-training (full). The results on Smarthome~\cite{Das_2019_ICCV} are shown in Tab.~\ref{tab1}. We test the 2s-AGCN~\cite{2sagcn2019cvpr} with the original 2D and 3D pose data extracted from LCRNet++~\cite{RogezWS18}, and 2D pose refined by our SSTA-PRS including or excluding the self-training step. Note that the 3D refined pose data is obtained by VideoPose3D~\cite{pavllo:videopose3d:2019} over the corresponding 2D poses. It shows that both components can improve the performance on action classification ($+1\sim2\%$ brought by SST-A and $+2\sim5\%$ by self-training). In addition, we show in Tab.~\ref{tab2} the results for two training modes, with either weak or strong supervision (i.e. with or without joint positions). We find that the results under strong supervision are below the ones using SST-A only. It suggests that the refined joint positions are too noisy to improve the regression branch. It shows also that
SST-A significantly improves the upstream localization and classification performance of SSTA-PRS, which is sufficient to improve action classification.\par
\begin{table}[ht]
\centering

\begin{center}
\scalebox{0.8}{
\setlength{\tabcolsep}{4.2mm}{
\begin{tabular}{ l l c  c }
\hline
\multirow{2}*{\textbf{2s-AGCN-Joint~\cite{2sagcn2019cvpr}}}& &\multicolumn{2}{c}{\textbf{Smarthome}}\\&
&\text{CS(\%)} &\text{CV2(\%)} \\
\hline
\hline

\text{  +2D-original~\cite{Das_2019_ICCV}}& &\text{52.9} &\text{47.5} \\
\text{  +2D SSTA-PRS (SST-A only)}& &\text{53.5} &\text{47.9} \\
\textbf{    +2D SSTA-PRS (full)}& &\textbf{55.7} &\textbf{49.0} \\
\hline
\text{  +3D-original~\cite{Das_2019_ICCV}}& &\text{49.1} &\text{49.6} \\
\text{  +3D SSTA-PRS (SST-A only)}& &\text{51.2} &\text{50.6} \\
\textbf{    +3D SSTA-PRS (full)}& &\textbf{54.0} &\textbf{53.3} \\

\hline

\end{tabular}
}}
\end{center}
\vspace{-0.2cm}
\caption{Mean per-class accuracy on Smarthome dataset using 2s-AGCN-Joint~\cite{2sagcn2019cvpr} with the skeleton data obtained by LCRNet++~\cite{RogezWS18} (original), SSTA-PRS using SST-A only (Sec.~\ref{ssta}) and using both SST-A and self-training(full) (Sec.~\ref{st}).}
\vspace{-0.2cm}
\label{tab1}
\end{table}

\begin{table}[ht]
\centering
\begin{center}
\scalebox{0.8}{
\setlength{\tabcolsep}{5.6mm}{
\begin{tabular}{ l c }
\hline
\textbf{SSTA-PRS Training Methods} &\textbf{Smarthome} CS(\%) \\
\hline
\hline
\text{Strong-supervision} &\text{51.4} \\
\textbf{Weak-supervision} &\textbf{54.0} \\
\hline
\end{tabular}
}
}
\end{center}
\vspace{-0.2cm}
\caption{Mean per-class accuracy on Smarthome dataset using the 3D skeleton data obtained by SSTA-PRS (2D) and VideoPose~\cite{pavllo:videopose3d:2019} (3D reconstruction over 2D skeleton)}
\label{tab2}
\vspace{-0.2cm}
\end{table}  


\subsection{Comparison With State-of-the-art}
\textbf{Toyota Smarthome:} we compare the final model, 2s-AGCN+SSTA-PRS, with the state-of-the-art methods on the Toyota Smarthome dataset. The results shown in Tab.~\ref{tab4} demonstrate that high-quality pose is a good modality for real-world action classification and our method achieves state-of-the-art performance with a large margin ($+3.8\%$ on CS and on CV2) over the Pose-based methods. Moreover, we apply our module for VPN~\cite{das2020vpn} as VPN+SSTA-PRS to show that the refined poses can also improve RGB-and-pose methods ($+4.4\%$ on CS and $+0.6\%$ on CV2).\par

\begin{table}[ht]
\centering

\begin{center}

\scalebox{0.75}{
\setlength{\tabcolsep}{1.6mm}{
\begin{tabular}{  l c c c c c}

\hline
\multirow{2}*{\textbf{Methods}}
&\multirow{2}*{\textbf{RGB}}
&\multirow{2}*{\textbf{Pose}}
&\multicolumn{3}{c}{\textbf{Smarthome}}\\
& & &\text{CS(\%)} &\text{CV1(\%)} &\text{CV2(\%)} \\
\hline
\hline
DT~\cite{Wang2011ActionRB}&\checkmark&$\times$&41.9&20.9&23.7\\ 
I3D~\cite{Carreira_2017_CVPR}&\checkmark&$\times$&53.4 &34.9& 45.1\\
I3D+NL~\cite{Wang2018NonlocalNN}  &\checkmark&$\times$& 53.6 &34.3& 43.9\\
AssembleNet++~\cite{Ryoo2020AssembleNetAM}(+obj)&\checkmark&$\times$& \textbf{63.6} &-& -\\

\hline
P-I3D~\cite{Das2019WhereTF} &\checkmark &\checkmark&54.2 &35.1 &50.3\\
Separable STA~\cite{Das_2019_ICCV}& \checkmark&\checkmark& 54.2& 35.2& 50.3\\
VPN~\cite{das2020vpn} &\checkmark &\checkmark&\text{60.8} &\textbf{43.8} &\text{53.5}\\
\textbf{VPN+PRS(ours)} &\checkmark &\checkmark&\textbf{65.2} &\textbf{-} &\textbf{54.1}\\
\hline

LSTM~\cite{Mahasseni2016RegularizingLS}&$\times$&\textbf{\checkmark}& 42.5 &13.4& 17.2\\

MS-AAGCN~\cite{shi_skeleton-based_2019}&$\times$&\textbf{\checkmark}&56.5 & - & -\\

2s-AGCN~\cite{2sagcn2019cvpr}&$\times$&\textbf{\checkmark}&57.1 & 22.1 & 49.7\\
\textbf{2s-AGCN+PRS(ours)}&$\times$&\text{\checkmark}&\textbf{60.9}& \textbf{22.5} &\textbf{53.5}\\
\textbf{5C-AGCN+PRS(ours)}&$\times$&\textbf{\checkmark}&\textbf{62.1}&\textbf{22.8} &\textbf{54.0}\\
\hline

\end{tabular}
}}

\end{center}
\vspace{-0.2cm}
\caption{Mean per-class accuracy comparison against state-of-the-art methods on the Toyota Smarthome dataset.}
\vspace{-0.2cm}
\label{tab4}
\end{table}

\textbf{Charades:} we are the first to use only skeleton data on Charades, so we only compare the frame-based mAP of the same baseline methods using different poses as input to validate the effectiveness of the pose estimation algorithm.
In this work, we use three types of input: 1) LCRNet3D poses extracted from LCRNet++~\cite{RogezWS18},
2)~LCRNet2D+VideoPose3D which uses video-based pose estimation system VideoPose3D~\cite{pavllo:videopose3d:2019} to reconstruct the LCRNet2D poses to 3D poses, 3)~SSTA-PRS 2D poses refined by our proposed algorithm.
We utilize three baseline methods to evaluate these poses: Bidirectional-LSTM~\cite{graves2005framewise}, Dilated-TCN~\cite{lea2017temporal} and TGM~\cite{TGM}. 
As shown in Tab.~\ref{tab:charades_table}, SSTA-PRS is consistently better than image-based LCRNet3D poses and video-based LCRNet2D+VideoPose3D poses for these three baselines (up to $+1.7\%, +0.6\%$ \wrt LCRNet3D and LCRNet2D+VideoPose3D respectively).\par
\begin{table}[ht]
\centering
\begin{center}

\scalebox{0.8}{
\setlength{\tabcolsep}{1.8mm}{

\begin{tabular}{lccc}
\hline
\multirow{2}*{\textbf{Methods}} & \multicolumn{3}{c}{\textbf{Charades}}\\

                     & Bi-LSTM~\cite{graves2005framewise} & Dilated-TCN~\cite{lea2017temporal} & TGM~\cite{TGM} \\ \hline \hline
LCRNet~3D~\cite{RogezWS18}            & 17.6              & 18.0   &18.9 \\
VideoPose~\cite{pavllo:videopose3d:2019} & 18.9            & 18.7    &20.1\\ \hline
\textbf{SSTA-PRS (ours)}    & \textbf{19.5} & \textbf{19.5}    &\textbf{20.6}\\ \hline
\end{tabular}
}
}
\end{center}
\vspace{-0.2cm}
\caption{Action detection performance on Charades using frame-based mAP.}
\vspace{-0.2cm}
\label{tab:charades_table}
\end{table}

\textbf{Kinetics-50:} we compare the final model with the state-of-the-art skeleton-based action recognition methods on the Kinetics-50 dataset (Tab.~\ref{tab_kinetics}). The methods used for comparison are the graph-based methods~\cite{2sagcn2019cvpr, shi_skeleton-based_2019} using OpenPose~\cite{OpenPose} skeleton data extracted by Yan et al.~\cite{Yan2018SpatialTG}. Our model based on SSTA-PRS refined poses achieves state-of-the-art performance ($44.7\%$ on Top1 and $73.1\%$ on Top5).

\begin{table}[ht]
\centering
\begin{center}
\scalebox{0.8}{

\setlength{\tabcolsep}{5.0mm}{
\begin{tabular}{ l c c }
\hline
\multirow{2}*{\textbf{Methods}} &\multicolumn{2}{c}{\textbf{Kinetics-50}} \\

&\text{Top-1(\%)} &\text{Top-5(\%)} \\
\hline
\hline
2s-AGCN~\cite{2sagcn2019cvpr}&40.0 &71.8 \\
MS-AAGCN~\cite{shi_skeleton-based_2019}&41.0 &72.4 \\
\hline
\textbf{2s-AGCN+SSTA-PRS (ours)} &\textbf{44.7} &\textbf{73.1} \\
\hline

\end{tabular}
}
}
\end{center}
\vspace{-0.2cm}
\caption{Classification accuracy comparison against state-of-the-art methods on the Kinetics-50 dataset.}
\vspace{-0.4cm}
\label{tab_kinetics}
\end{table}

\section{Conclusion}

In this work, we propose a novel method to extract pose sequences from challenging real-world videos. Owing to the proposed novel aggregation mechanism (SST-A) and weakly-supervised self-training framework, our method can be applied on 1) videos in low-resolution and 2) videos containing human body occlusions and truncations. When applied on a real-world activity recognition dataset (\eg Smarthome), the poses extracted from our model can be used to achieve the best results which show the effectiveness of our method. However, it still has limitations since we mainly focus on improving the quality of poses. For the tasks concerning human-object interactions (\eg Charades), our method does not bring too much boosting. One of the future directions could be combining our method with RGB-based methods together to further improve the recognition results.
\paragraph{Acknowledgement} This work is supported by Toyota Motor Europe (TME). We are grateful to Sophia Antipolis - Mediterranean ”NEF” computation cluster for providing resources and support.

{\small
\bibliographystyle{ieee_fullname}
\bibliography{egbib}
}

\end{document}